\documentclass[runningheads]{llncs}
\usepackage{graphicx}

\usepackage{amssymb}
\usepackage{amsmath}
\usepackage{lipsum}
\usepackage{hyperref}
\usepackage{multirow}
\usepackage{xcolor}
\usepackage{enumitem}
\usepackage{bm}
\usepackage{color}
\usepackage{makecell}

\usepackage{amssymb}
\usepackage{booktabs}
\usepackage{float}
\begin{document}
\title{Bridging Modality Gap for Visual Grounding with Effecitve Cross-modal Distillation}
\titlerunning{Cross-modal Distillation for Visual Grounding}
%
\author{Jiaxi Wang\inst{1} \and
Wenhui Hu\inst{1} \and
Xueyang Liu\inst{1} \and Beihu Wu\inst{1} \and Yuting Qiu\inst{1} \and YingYing Cai\inst{1}}

\authorrunning{J. Wang et al.}

\institute{\textsuperscript{1}Peking University, No.5 Yiheyuan Road Haidian District, Beijing, China }

\maketitle              
\begin{abstract}
Visual grounding aims to align visual information of specific regions of images with corresponding natural language expressions. Current visual grounding methods leverage pre-trained visual and language backbones independently to obtain visual features and linguistic features. Although these two types of features are then fused through elaborately designed networks, the heterogeneity of the features renders them unsuitable for multi-modal reasoning. This problem arises from the domain gap between the single-modal pre-training backbones used in current visual grounding methods, which can hardly be bridged by the traditional end-to-end training method. To alleviate this, our work proposes an Empowering Pre-trained Model for Visual Grounding (EpmVG) framework, which distills a multimodal pre-trained model to guide the visual grounding task. EpmVG relies on a novel cross-modal distillation mechanism that can effectively introduce the consistency information of images and texts from the pre-trained model, reducing the domain gap in the backbone networks, and thereby improving the performance of the model in the visual grounding task. Extensive experiments have been conducted on five conventionally used datasets, and the results demonstrate that our method achieves better performance than state-of-the-art methods.

\end{abstract}
%
%
\section{Introduction}
\label{introduction}
Visual grounding~\cite{mao2016generation,plummer2015flickr30k}, a key research area at the intersection of computer vision~\cite{stockman2001computer} and natural language processing~\cite{chowdhary2020natural}, aims to link natural language descriptions to visual content by pinpointing the specific area of an image described by the language. Unlike traditional object detection~\cite{redmon2018yolov3,ren2015faster} methods that can only identify limited categories in visual training data, visual grounding can detect novel categories and attributes expressed in free-form texts. This has garnered significant attention in recent years and has widespread applications like object search~\cite{hu2016natural,li2017deep}, video analysis~\cite{li2023g2l,zhang2022unsupervised}, automation~\cite{groover2001automation}, robot navigation~\cite{gul2019comprehensive}, and augmented reality~\cite{craig2013understanding}.

\begin{figure}[t] 
	\centering 
	\includegraphics[width=1\textwidth, angle=0]{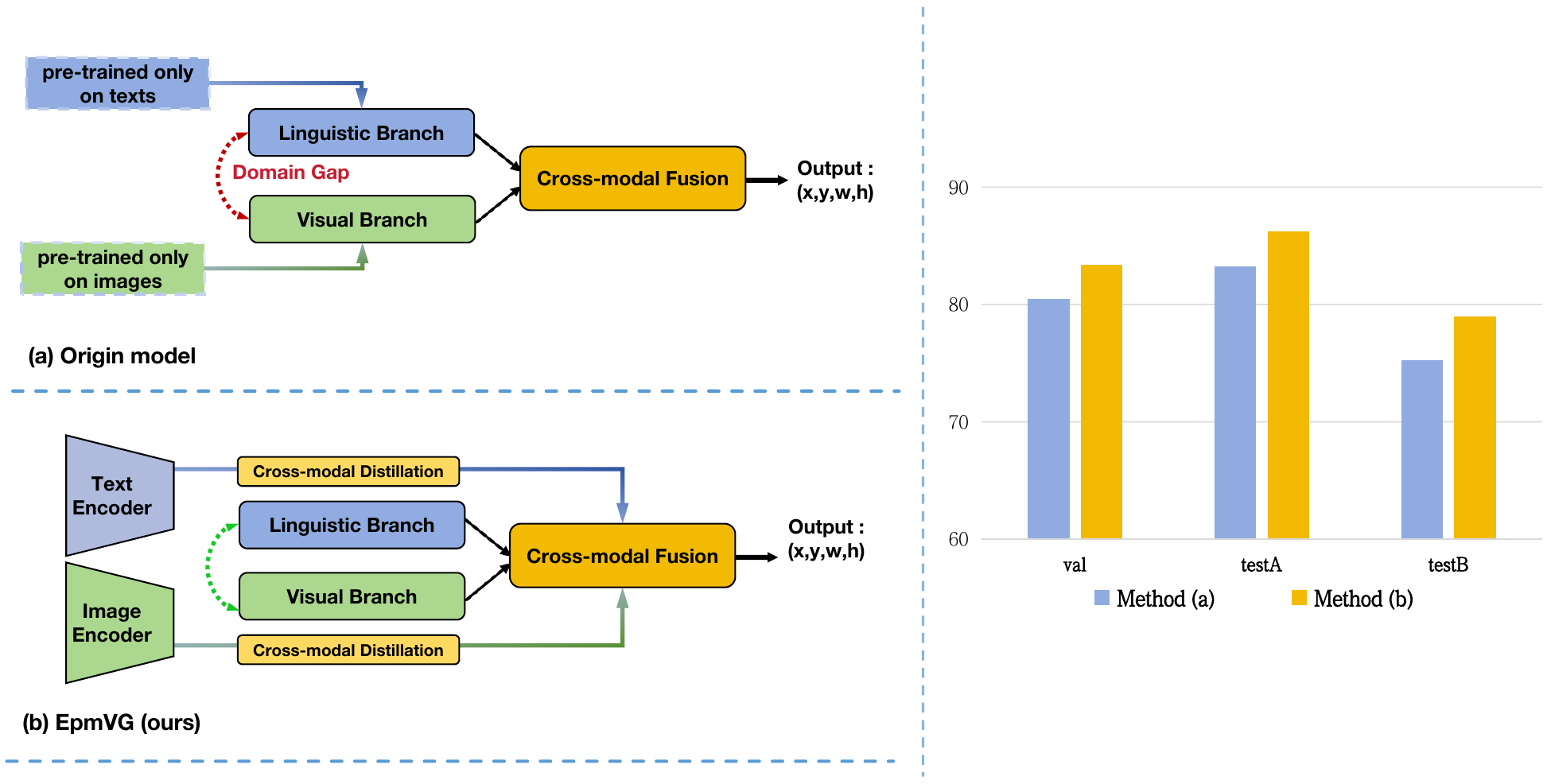}	
	\caption{\textbf{Left:} (a) Basic visual grounding model and (b) our proposed EpmVG framework. \textbf{Right:} The comparison of two methods on the RefCOCO~\cite{yu2016modeling} dataset.}
	\label{fig_mom0}%
\end{figure}

The early methods of visual grounding primarily focused on extending the application of one-stage and two-stage object detection architectures. One-stage methods~\cite{ye2022shifting,yang2022improving,deng2021transvg,huang2021look} utilize a pre-trained fully convolutional network (e.g., Darknet53~\cite{redmon2018yolov3}, ResNet~\cite{he2016deep}) to directly generate pixel-level feature maps and return the most likely candidate for the query text using manually defined dense anchors. These methods are simple and efficient for learning or inference, but they struggle with complex queries that involve multiple objects and relationships. Two-stage methods~\cite{mao2016generation,nagaraja2016modeling,wang2018learning,yu2016modeling} first generate a set of sparse region proposals and then exploit region-expression matching to find the best one. In these two-stage methods, modular attention networks~\cite{yu2018mattnet}, various graphs~\cite{yang2020graph}, and multimodal trees~\cite{liu2019improving} are designed to better model multimodal relationships. However, the complex fusion modules cannot be jointly learned with the detectors, which may limit their ability in multimodal reasoning. Recently, the visual grounding methods~\cite{ye2022shifting,yang2022improving,deng2021transvg} based on the Transformer~\cite{vaswani2017attention} structure have been introduced into the field of visual localization due to their strong generalization ability and faster speed. Most of the current state-of-the-art methods are based on the Transformer architecture. In this paper, we focus on designing a better Transformer-based visual grounding method.

Despite the success achieved by the existing Transformer-based visual grounding methods~\cite{ye2022shifting,yang2022improving,deng2021transvg}, we argue that they suffer from the \textbf{modality domain gap}. Specifically, during the pre-training stage, both the text and image encoder are pre-trained on their respective single-modal datasets. However, as shown in Figure \ref{fig_mom0} left(a), visual grounding is naturally a cross-modal task, and this domain gap may lead to inferior results.


In this paper, we propose an \textbf{E}mpowering \textbf{p}re-trained \textbf{m}odel for \textbf{V}isual \textbf{G}rounding (EpmVG) to address the domain gap issue. The proposed EpmVG can transfer the cross-modal correlation information contained in C\-LIP~\cite{radford2021learning} by a novel \textbf{C}ross-modal \textbf{D}istillation loss (CD). We separately use the frozen CLIP~\cite{radford2021learning} model's visual encoder and text encoder to generate soft labels, which are used to constrain the corresponding visual branch and linguistic branch. The experimental results shown in Figure~\ref{fig_mom0} right(b) are better than the previous methods, indicate that it can reduce the modality domain gap between images and texts, facilitates cross-modal alignment between queries and related areas. We evaluate our framework on five widely used visual grounding datasets, including ReferItGame~\cite{kazemzadeh2014referitgame}, Flickr30K Entities~\cite{plummer2015flickr30k}, RefCOCO~\cite{yu2016modeling}, RefCOCO+~\cite{yu2016modeling}, RefCOCOg~\cite{mao2016generation}, and our method effectively improves the performance of the original visual grounding method. Note that the proposed EpmVG can be easily applied to other visual grounding models. We use QRNet~\cite{ye2022shifting} as our visual grounding model of EpmVG, which also significantly achieves absolute improvements over the original method and outperforms other existing methods.

The main contributions of this article are threefold:
\begin{itemize}
    \item We analyze a problem existing in the current visual grounding tasks. Due to the domain gap issue in the pre-training stage, the visual and language backbones lack cross-modal correlation information between images and texts.
    \item We propose a framework EpmVG  to transfer the correlation information of images and text to the visual grounding model through image-text joint distillation to improve the performance of the model.
    \item Extensive experiments are conducted to verify the advantages of our approach, and significantly improved results are shown on several popular benchmarks.
\end{itemize}

\section{Related Work}
\subsection{Visual Grounding}
The current visual grounding methods can be roughly classified into two categories, namely two-stage methods \cite{wang2019neighbourhood,cao2022deep,chen2021ref,ye2022shifting} and one-stage methods \cite{ye2021one,deng2021transvg,kamath2021mdetr,cao2022locvtp}.
\noindent \textbf{Two-stage methods.} The first stage involves generations of proposals, which are candidate regions containing objects in images that can be selected in the next stage. In the second stage, the matching similarities between the region proposal and text expressions will be measured and leveraged to perform a ranking task. The best-matched proposal thus can be found. Some works focus on the second stage to improve the model performance. MAttNet \cite{yu2018mattnet} and RvG-Tree \cite{hong2019learning} draw on the logical or structural decomposition of language expression to better reason the fine-grained linguistic information, to more accurately construct alignment with visual features. However,  the features proposed by the first stage may not contain the information the second stage needs, and such a gap makes the optimization in the second stage futile.  To solve this, QRNet \cite{ye2022shifting} further alleviates this problem by replacing the query-agnostic visual module with a query-aware module, thus providing more representative features for multimodal reasoning.

\noindent \textbf{One-stage methods.}
The one-stage method aims to overcome the weakness of the two-stage method by eliminating the possibility of the malfunctions in first stage. By fusing expression features with visual features, it can directly predict target objects position with bounding boxes. Yang et al. \cite{yang2019fast} embed text query to a 768D vector with a BERT \cite{devlin2018bert} model and extract visual features with Darknet-53 \cite{redmon2018yolov3}, and then they perform feature fusion by broadcasting and concatenating text features with visual and spatial features. Yang et al. \cite{yang2020improving} proposes to reduce the ambiguity of reference by identifying sub-query information, instead of using single vector representation to embed the entire sentences. To improve cross-modal understanding of modal, Ye et al. \cite{ye2021one} use three filters of three dimensions to filter visual features in corresponding to different facets of a text query, which are obtained by taking structural and contextual information into account. 

\noindent \textbf{Transformer-based Method.} Dosovitskiy et al.'s pioneer ViT \cite{dosovitskiy2020image} shows that pure transformer frameworks can effectively perform image classification tasks. Recent works like TTSR by Yang et al. \cite{yang2020learning} and Deformable DETR by Zhu et al. \cite{Zhu2020Deformable} use transformers for specific vision tasks like image super-resolution and small object detection. To improve the performance of ViT, Liu et al. \cite{liu2021Swin} proposed a new hierarchical ViT called Swin transformer. These transformer-based visual backbones provide support for visual grounding models. TransVG \cite{deng2021transvg} and MDETR \cite{kamath2021mdetr} both draw on the transformer to solve the problem of overfitting in certain scenarios and a long tail of open and free textual expressions respectively. QRNet \cite{ye2022shifting} employs a Swin encoder to extract visual features and proposes a fine-grained query modulation visual grounding framework.

\subsection{Knowledge Distillation} 

Knowledge Distillation \cite{Gou2021knowledge} aims at scaling down large models with huge amounts of parameters, to make it possible to achieve similar task performance with fewer resources. It usually involves training a student model by utilizing the knowledge of a teacher model. Some work \cite{Abnar2021trans,cao2023iterative} studies how to improve model performance by using different ways of distillation. To facilitate the application of ViT, Touvron et al. \cite{Touvron2020} take the distillation method, which involves training a student transformer by a powerful teacher image classifier, thus scaling the training work down with less data. In order to address the lack of large-scale training data for open-vocabulary text-visual tasks, ViLD \cite{Gu2021open} performs distillation by using pre-trained open-vocabulary image classification models, like CLIP~\cite{radford2021learning} and ALIGN \cite{Jia2021scaling}, as a teacher and modifying a vanilla detector as a student. Its detection categories are up to more than 1000 compared with tens of that of previous methods.

\section{Method}
\begin{figure}[t]
	\centering 
	\includegraphics[width=1\textwidth, angle=0]{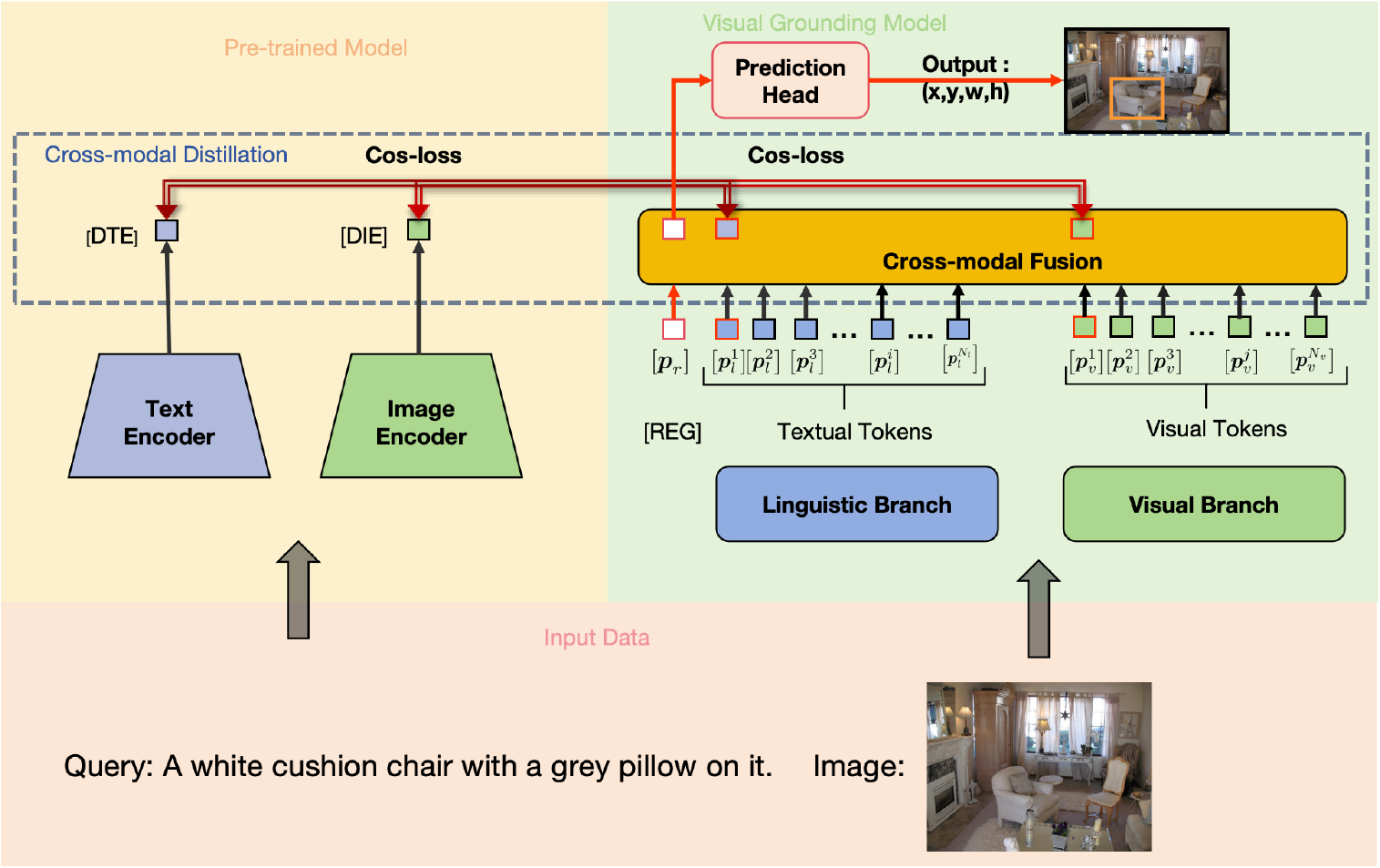}
	\caption{An overview of EpmVG framework. It consists of two main components: (1) a visual grounding model; and (2) a distillation module.}
	\label{fig_mom1}%
\end{figure}

In this work, we propose to transfer the visual and linguistic knowledge contained in the pre-trained model into a visual grounding model, namely \textbf{E}mpowering \textbf{P}re-trained \textbf{M}odel for \textbf{V}isual \textbf{G}rounding (EpmVG). The EpmVG structure is shown in Figure~\ref{fig_mom1}. There are two main components in EpmVG: (1) a visual grounding model, and (2) a distillation module. In the following subsections, we first formulate the visual grounding task and then introduce the architecture of EpmVG.
\subsection{Visual Grounding Module} 

The main purpose of the visual grounding task is to map a query to a specific region of the image. This query corresponds to an object in the image, which can be a sentence or a phrase. The task can be defined as follows: given an image $I$ and a query $q$, the model needs to predict a bounding box $\mathbf{b}=\{x, y, w, h\}$ that accurately surrounds the target object described by the query.

As shown in Figure~\ref{fig_mom1}, our visual grounding model is designed based on a typical end-to-end visual grounding architecture TransVG~\cite{deng2021transvg}. There are four main components in our visual grounding model: (1) a visual branch, (2) a linguistic branch, (3) a cross-modal fusion, and (4) a prediction module.

\textbf{Visual Branch.} The vision branch starts with a convolutional backbone network, followed by a vision transformer. We utilize the commonly used ResNet~\cite{he2016deep} as the backbone network. The vision transformer is a stack of transformer encoder layers. Each transformer encoder layer consists of a multi-head self-attention layer and an FFN.

First, the visual branch input is an image $\mathbf{z}_0 \in \mathbb{R}^{3 \times H_0 \times W_0}$, and then the backbone network is used to generate a two-dimensional feature map $\mathbf{z} \in \mathbb{R}^{C \times H \times W}$.

Then, this branch uses the 1×1 convolutional layer to reduce the dimension of channels of $\mathbf{z}$, and obtain $\mathbf{z}^{\prime} \in \mathbb{R}^{C_v \times H \times W}$. In order to conform to the input format of the transformer encoding layer, $\mathbf{z}^{\prime}$ is flattened to $\mathbf{z}_v \in \mathbb{R}^{C_v \times N_v}$, where $N_v=H \times W$ is the number of input tokens. Finally, the visual transformer extracts the visual context features and outputs the visual embedding $\mathbf{f}_v$, which has the same shape as $\mathbf{z}_v$.

\textbf{Linguistic Branch.} 
We utilize the linguistic branch, typically implemented as a pre-trained BERT~\cite{devlin2018bert} model, to encode the query sentence. The language embedding is denoted as $\mathbf{f}_l \in \mathbb{R}^{C_l \times N_l}$, where $N_l$ is the number of language tokens.

\textbf{Cross-modal Fusion.} The core structure of this module is a visual-linguistic transformer. To make the feature embeddings of different modalities have the same dimensions, we apply a linear projection layer to map them into embeddings with the same channel dimensions $\mathbb{R}^D$. The projected visual and linguistic features are denoted by $\mathbf{p}_v \in \mathbb{R}^{D \times N_v}$ and $\mathbf{p}_l \in \mathbb{R}^{D \times N_l}$. Then, $\mathbf{p}_v$ and $\mathbf{p}_l$ are connected by inserting a learnable embedding (namely a [REG] token) at the beginning of the connection sequence, and formulate the joint input tokens of the visual-linguistic transformer as: 
\begin{equation}
\mathbf{X}=[\mathbf{p}_r, \mathbf{p}_v, \mathbf{p}_l],
\end{equation} 
\begin{equation}
    \mathbf{p}_v=[\mathbf{p}_v^1, \mathbf{p}_v^2, \cdots, \mathbf{p}_v^{N_v}],
\end{equation}
\begin{equation}
\mathbf{p}_l=[\mathbf{p}_l^1, \mathbf{p}_l^2, \cdots, \mathbf{p}_l^{N_l}],
\end{equation}
where $\mathbf{p}_r$ represents the learnable embedding of [REG] token, $\mathbf{p}_v$ represents the visual tokens and $\mathbf{p}_l$ represents the linguistic tokens. $\mathbf{p}_l^1$ is the [CLS] token’s representation which is regarded as the contextual textual feature. And $\mathbf{p}_v^1$ is denoted by [VLS] which is regarded as the global visual feature.

After obtaining the input $\mathbf{X} \in \mathbb{R}^{D \times\left(N_v+N_l+1\right)}$ in the joint embedding space as mentioned above, we apply a multi-layer visual-linguistic transformer to perform intra- and inter-modal reasoning on the joint sequence through an attention mechanism. Finally, we obtain the output of this module $\mathbf{X^{\prime}}=[\mathbf{p^{\prime}}_r, \mathbf{p^{\prime}}_v^1, \mathbf{p^{\prime}}_v^2,  \cdots,  \newline \mathbf{p^{\prime}}_v^{N_v}, \mathbf{p^{\prime}}_l^1, \mathbf{p^{\prime}}_l^2, \cdots, \mathbf{p^{\prime}}_l^{N_l}]  \in \mathbb{R}^{D \times\left(N_v+N_l+1\right)}$.

\textbf{Prediction Module.} Finally, a prediction module is used to handle the output representation of the [REG] token from the Cross-modal Fusion module. The primary task of this prediction module is to predict the coordinates of the bounding box $\mathbf{b}$ based on this output representation. In short, the computing process is formulated as follows: 
\begin{equation}
\begin{aligned}
\mathbf{p^{\prime}}_r & =\operatorname{ReLU}\left(\operatorname{MLP}\left(\operatorname{ReLU}\left(\text { MLP }\left(\mathbf{p^{\prime}}_r\right)\right)\right)\right), \\
\mathbf{p^{\prime}}_r & =\operatorname{MLP}\left(\mathbf{p^{\prime}}_r\right), \\
\mathbf{b} & =\operatorname{Sigmoid}\left(\mathbf{p^{\prime}}_r\right) .
\end{aligned}
\end{equation}

\subsection{Cross-modal Distillation Module}
In this section, we introduce the proposed framework of EpmVG. We utilize CLIP~\cite{radford2021learning} as our pre-trained model. Because the CLIP~\cite{radford2021learning} model uses a large number of image and text pairs for training, the model can understand the semantic relationship between images and text. For example, given an image and a text describing the image, a CLIP~\cite{radford2021learning} model can understand the relationship between the two and apply this understanding to new images and text. The visual grounding task relies heavily on the model's understanding of the semantic relationship between images and text. In the EpmVG, we utilize Cross-modal Distillation loss  to effectively transfer the pre-trained model's understanding of the semantic relationship between images and text to the visual grounding model, the details of which are shown as follows.

For a given image $\mathbf{z}_0 \in \mathbb{R}^{3 \times H_0 \times W_0}$, we utilize CLIP~\cite{radford2021learning}’s image encoder to extract one-dimensional features $\hat{\mathbf{p}}_{cv} \in \mathbb{R}^{d}$ that incorporate visual semantic information and corresponding text relationship information, where $\hat{\mathbf{p}}_{cv}$ is named [DIE]. To the same, for a given query sentence, we utilize CLIP~\cite{radford2021learning}’s text encoder to extract one-dimensional features $\hat{\mathbf{p}}_{cl} \in \mathbb{R}^{d}$ that incorporate textual semantic information and corresponding visual relationship information, where $\hat{\mathbf{p}}_{cl}$ is named [DTE]. Then we utilize $\hat{\mathbf{p}}_{cv}$ and $\hat{\mathbf{p}}_{cl}$ to guide at the beginning positions of the tokenized image expression $\mathbf{p^{\prime}}_v^1$ and at the beginning positions of the tokenized language expression  $\mathbf{p^{\prime}}_l^1$ to better acquire the knowledge of the semantic relationship between images and text in the pre-trained model. The distillation process is formulated as:

\begin{equation}
\hat{\mathbf{P}}_{cv}=\mathcal{\operatorname{AvgPool}}(\hat{\mathbf{p}}_{cv}, \mathbf{p^{\prime}}_v^1),
\end{equation}
\begin{equation}
\hat{\mathbf{P}}_{cl}=\mathcal{\operatorname{AvgPool}}(\hat{\mathbf{p}}_{cl}, \mathbf{p^{\prime}}_l^1),
\end{equation}
We utilize Adaptive Average Pooling~\cite{van2019evolutionary} to transform $\hat{\mathbf{p}}_{cv}$ into the same dimension as $\mathbf{p^{\prime}}_v^1$. The same operation is also applied to $\hat{\mathbf{p}}_{cl}$. Where $\hat{\mathbf{P}}_{cv}\in \mathbb{R}^{D}$  and $\hat{\mathbf{P}}_{cl}\in \mathbb{R}^{D}$ are the feature(s) after transformation.
\begin{equation}
\mathcal{L}_{\text {distillation }}=\alpha \cdot \mathcal{L}_{\text {cos }}(\mathbf{p^{\prime}}_v^1, \hat{\mathbf{P}}_{cv})+\beta \cdot \mathcal{L}_{\text {cos }}(\mathbf{p^{\prime}}_l^1, \hat{\mathbf{P}}_{cl}),
\end{equation}
where $\mathcal{L}_{\text {cos }}(\cdot)$ is the cosine loss~\cite{wang2018cosface}, $\alpha$ and $\beta$ are the distillation weights for balancing the two modalities.

Finally, we get both the prediction module’s prediction $\mathbf{b}=(x, y, w, h)$, and the normalized ground-truth box as $\hat{\mathbf{b}}=(\hat{x}, \hat{y}, \hat{w}, \hat{h})$. And the training objective of our EpmVG is:
\begin{equation}
\mathcal{L}=\mathcal{L}_{\text {smooth-11 }}(\mathbf{b}, \hat{\mathbf{b}})+\lambda \cdot \mathcal{L}_{\text {giou }}(\mathbf{b}, \hat{\mathbf{b}})+\mathcal{L}_{\text {distillation}},
\end{equation}
where $\mathcal{L}_{\text {smooth-11 }}(\cdot)$ and $\mathcal{L}_{\text {giou }}(\cdot)$ are the smooth L1 loss~\cite{girshick2015fast} and GIoU loss~\cite{rezatofighi2019generalized}, respectively. $\lambda$ is the weight coefficient of GIoU loss to balance these losses.


\section{Experiments}
\subsection{Datasets and Evaluation}

\textbf{Datasets.} We conduct experimental validation on five common public datasets, ReferItGame \cite{kazemzadeh2014referitgame}, Flickr30k Entities \cite{plummer2015flickr30k}, RefCOCO \cite{yu2016modeling}, RefCOCO+ \cite{yu2016modeling} and RefCOCOg \cite{mao2016generation}.

\textbf{Evaluation Metric.} We adopt the same evaluation metrics as~\cite{deng2021transvg}. Specifically, if the Intersection over Union (IoU) of the predicted bounding box and the actual (ground-truth) bounding box exceeds 0.5, then this prediction is considered to be correct.

\subsection{Implementation Details}

The dimensions of the input image are established at $H_o=640$ and $W_o=640$. In the process of resizing the image, the original aspect ratio is maintained. The image's longer side is adjusted to 640, and the shorter side is padded to 640 using the average value of the RGB channels. Meanwhile, for language queries, we set the maximum length to 40 for RefCOCOg, and to 20 for other datasets.

During training, we use the CLIP~\cite{radford2021learning} model as the pre-trained model component of our EpmVG framework. We adhere to the procedures outlined in TransVG \cite{deng2021transvg} for processing input images and sentences. Similarly, we adopt the training configuration utilized in TransVG, which employs an AdamW optimizer with a weight decay of $10^{-4}$, establishes a batch size of 64, and sets a dropout ratio of 0.1 for the FFN in the Transformer. The learning rate for pre-trained parameters in the visual and linguistic branch is set at $10^{-5}$, while it is set at $10^{-4}$ for the remaining parameters. The parameters without pre-training are randomly initialized with Xavier. We train our model 180 epochs (with a corresponding learning rate drop at 120) for RefCOCO+, and 90 epochs (with a corresponding learning rate drop at 60) for other datasets. The weight coefficient $\gamma$ is set to 1. The distillation weight coefficients $\alpha$ and $\beta$ are set to 1.5 and 2. We also adopt the data augmentation strategies commonly employed in \cite{liao2020real,yang2020improving,yang2019fast}.

\subsection{Ablation Study }

In this section, we conducted a series of ablation experiments to validate the effectiveness of our proposed framework and each of its components. Initially, we selected different methods as our visual grounding module and compared the performance of the model before and after distillation on five widely used public datasets to verify the effectiveness of our proposed framework.

\textbf{Effectiveness of the Overall Framework.} The overall effectiveness of the framework was demonstrated by constructing the visual grounding part of EpmVG based on TransVG~\cite{deng2021transvg} and QRNet~\cite{ye2022shifting} methods and comparing the results before and after distillation on five public datasets. As shown in Tables \ref{tab1} and \ref{tab2}, our method significantly outperformed the original methods on all five datasets after adding cross-modal distillation module, effectively validating the efficacy of our framework.

\begin{table}[t]
\caption{Comparison with the method without distillation on the test set of ReferItGame~\cite{kazemzadeh2014referitgame} and Flickr30K~\cite{plummer2015flickr30k} Entities in terms of top-1 accuracy (\%).}

\small
\begin{center}
\scalebox{0.8}[0.8]{
\setlength
\tabcolsep{8pt}
    \begin{tabular}{c|c|c|c|c|c c c}
    \toprule 
        VG & \multirow{2}{*}{Backbone}  & \multirow{2}{*}{Distillation}& ReferItGame & Flickr30K & \multicolumn{3}{c}{RefCOCO} \\   method &  & & test & test & val & testA & testB \\
        \hline \hline 
        \multirow{2}{*}{TransVG} & \multirow{2}{*}{ResNet-50}  &w/o & 69.76 & 78.47 &80.49 &83.28 &75.24 \\
        &   &w/ & $\mathbf{7 0 . 6 5}$ & $\mathbf{7 9 . 2 9}$ &$\mathbf{83.39}$&$\mathbf{86.20}$&$\mathbf{78.98}$\\
        \hline 
        \multirow{2}{*}{TransVG} & \multirow{2}{*}{ResNet-101} &w/o & 70.73& 79.10 &80.83 &83.38 &76.94 \\
         &   &w/ & $\mathbf{7 1 . 1 6}$ & $\mathbf{79.58}$ &$\mathbf{82.49}$&$\mathbf{85.23}$&$\mathbf{79.51}$\\
    \hline
     \multirow{2}{*}{QRNet}& \multirow{2}{*}{Swin-S}& w/o & 74.61&81.95 &84.01&85.85&82.34\\
     &  & w/ & $\mathbf{75.84}$& $\mathbf{83.14}$&$\mathbf{87.04}$&$\mathbf{89.21}$&$\mathbf{82.90}$\\
     \bottomrule
    \end{tabular}
} 
\end{center}
\label{tab1}
\vspace{-1.0cm}
\end{table}

\begin{table}[t]
\caption{Comparisons with state-of-the-art methods on RefCOCO~\cite{yu2016modeling}, RefCOCO+~\cite{yu2016modeling} and RefCOCOg~\cite{mao2016generation} in terms of top-1 accuracy (\%). }

\small
\begin{center}
\scalebox{0.8}[0.8]{
\setlength
\tabcolsep{8pt}
    \begin{tabular}{c  |c   |c  |c c c  |c c c }
				\toprule 
				VG & \multirow{2}{*}{Backbone}  & \multirow{2}{*}{Distillation}& \multicolumn{3}{c|}{RefCOCO+} & \multicolumn{3}{c}{RefCOCOg} \\ 
				
				method &   & & val & testA & testB & val-g & val-u & test-u \\
				\hline\hline
				\multirow{2}{*}{TransVG} & \multirow{2}{*}{ResNet-50}  &w/o & 66.39 & 70.55 & 57.66 & 66.35 & 67.93 & 67.44 \\
				& &w/ & $\mathbf{69.67}$& $\mathbf{74.70}$& $\mathbf{59.87}$& $\mathbf{71.61}$& $\mathbf{71.92}$& $\mathbf{71.40}$\\
				\hline
 \multirow{2}{*}{TransVG} & \multirow{2}{*}{ResNet-101} &w/o & 68.00 & 72.46 & 59.24 & 68.03 & 68.71 &67.98 \\
				& &w/ & $\mathbf{69.34}$& $\mathbf{73.57}$& $\mathbf{60.85}$& $\mathbf{69.54}$& $\mathbf{71.02}$& $\mathbf{70.74}$\\
				\hline
         \multirow{2}{*}{QRNet}& \multirow{2}{*}{Swin-S}& w/o & 72.94& 76.17& 63.81& 71.89& 73.03&72.52\\
          & & w/ & $\mathbf{76.45}$& $\mathbf{81.16}$& $\mathbf{68.12}$& $\mathbf{76.15}$& $\mathbf{79.32}$&$\mathbf{76.61}$\\
    \bottomrule
    \end{tabular}
} 
\end{center}
\label{tab2}
\vspace{-0.5cm}
\end{table}

Next, we validated the effectiveness of each component of the framework on the RefCOCO~\cite{yu2016modeling} dataset, where the VG model was based on TransVG~\cite{deng2021transvg}, and the backbone network was ResNet-50.

\textbf{Distillation Loss.} As shown in Table \ref{tab3}, we studied the effectiveness of the distillation loss function and multimodal distillation in cross-modal distillation module. When cos loss was replaced with L1 loss, the performance decreased by 0.96\%, 1.39\%, and 1.55\% on val, testA, and testB, respectively, indicating that cos loss is more effective than L1 loss. When text modality distillation was removed, the performance decreased by 13.11\%, 25.28\%, and 24.65\%, respectively. When image modality distillation was removed, the performance decreased by 10.64\%, 23.83\%, and 23.75\%, respectively. We found that the distillation of both modalities is crucial, and only by distilling both text and images can we more effectively learn the semantic consistency information in the pre-trained model.

\begin{table}[t]
    \caption{Ablation studies of distillation loss function and distillation modes in EpmVG.}
    
    \begin{center}
	\scalebox{0.8}[0.8]{
	\setlength
	\tabcolsep{9.4pt}
           \begin{tabular}{c c cc ccc}
           \toprule
            \multirow{2}{*}{ Mode }& Distillation & Image & Text & \multicolumn{3}{c}{ RefCOCO } \\ 
            & Loss & Distillation& Distillation & val & testA & testB \\
            \hline\hline
            \#1& Cos.& \checkmark & \checkmark &$\mathbf{83.39}$ & $\mathbf{86.20}$ & $\mathbf{78.98}$ \\
            \hline
             \#2& L1& \checkmark & \checkmark & 82.43& 84.81&77.43\\ 
             \#3& Cos.& \checkmark & \texttimes& 70.28& 60.92&54.33\\
             \#4& Cos.& \texttimes& \checkmark & 72.75& 62.37&55.23\\ 
             \bottomrule
        \end{tabular}
		} 
	\end{center}
	\label{tab3}
	\vspace{-1.0cm}
\end{table}


\textbf{Pretrained Model Parameters.} We experimentally validated the impact of whether the parameters of the pre-training model CLIP~\cite{radford2021learning} are frozen during the distillation process on performance. As shown in Table \ref{tab5}, when the parameters were not frozen, the performance decreased by 0.36\%, 0.93\%, and 0.94\%, respectively. This suggests that if the CLIP~\cite{radford2021learning} model is altered during the distillation process, it may cause the knowledge learned by the VG model to change, thereby affecting its performance.

\begin{table}[t]
    \caption{Ablation studies of whether the pre-trained model is frozen in EpmVG.}
    
    \begin{center}
	\scalebox{0.8}[0.8]{
	\setlength
	\tabcolsep{9.4pt}
           \begin{tabular}{c c ccc}
           \toprule
            \multirow{2}{*}{ Mode }& Pretrained Model & \multicolumn{3}{c}{ RefCOCO } \\ 
            & Parameters & val & testA & testB \\
            \hline\hline
            \#1& frozen &$\mathbf{83.39}$ & $\mathbf{86.20}$ & $\mathbf{78.98}$ \\
            \hline
             \#2& unfrozen & 83.03& 85.27&78.04\\ 
             \bottomrule
        \end{tabular}
		} 
	\end{center}
	\label{tab5}
	\vspace{-0.5cm}
\end{table}

\subsection{Comparisons with State-of-the-art Methods}

We present the comparison results on ReferItGame~\cite{kazemzadeh2014referitgame} and Flickr30k~\cite{plummer2015flickr30k} Entities in Table \ref{tab6}. ReferItGame is a cooperative game that gathers queries by having one player create a referential phrase for a designated object, while another player is tasked with selecting the correct object. As for Flickr30k Entities, the queries are derived from the phrases found in the title. We observe that the proposed EpmVG surpasses previous work. The cross-modal distillation we designed effectively introduces the semantic consistency information of images and text in CLIP~\cite{radford2021learning} into the VG model. This solves the problem of the domain gap between the single-modal pre-training of the backbone network and the downstream multi-modal tasks in the VG model that we mentioned earlier, successfully improving accuracy.

\begin{table}[t]
	\caption{ The performance comparisons (Acc@0.5) on ReferItGame~\cite{kazemzadeh2014referitgame} and Flickr30K Entities~\cite{plummer2015flickr30k}.}
	\small
	\begin{center}
		\scalebox{0.8}[0.8]{
			\setlength
			\tabcolsep{9.4pt}
			\begin{tabular}{c  |c  |c  |c  }
				\toprule
				\multirow{2}{*}{Models} & \multirow{2}{*}{Backbone} & \multicolumn{1}{c|}{ReferItGame} & \multicolumn{1}{c}{Flickr30K}\\
				& & test & test\\
				\hline
				\hline
				\textbf{\textit{Two-stage:}}  &  &  &\\
				VC~\cite{zhang2018grounding} & VGG16 & 31.13 & - \\
				MAttNet~\cite{yu2018mattnet} &	ResNet-101  & 29.04 & - \\
				Similarity Net~\cite{wang2018learning}& ResNet-101 & 34.54 & 60.89\\
				DDPN~\cite{yu2018rethinking} & ResNet-101 &  63.00 & 73.30\\
                    DIGN~\cite{mu2021disentangled}& VGG-16 & 65.15 & 78.73 \\
				\hline
				\textbf{\textit{One-stage:}}  &  & &\\
				FAOA~\cite{yang2019fast} & DarkNet-53 & 60.67 & 68.71\\
				RCCF~\cite{liao2020real} & DLA-34 & 63.79 & -\\
				ReSC-Large~\cite{yang2020improving} & DarkNet-53 & 64.60 & 69.28\\
				TransVG~\cite{deng2021transvg} & ResNet-101 &  70.73 & 79.10 \\
                    TransVG (Swin) & Swin-S & 70.86 & 78.18 \\
                    QRNet~\cite{ye2022shifting} & Swin-S & 74.61 & 81.95 \\
                    \hline
                    EpmVG(ours) & Swin-S &  \textbf{75.84} & \textbf{83.14} \\
				
				\bottomrule
			\end{tabular}
		} 
	\end{center}
	\label{tab6}
	\vspace{-1.0cm}
\end{table}

\begin{table}[t]
	\caption{The performance comparisons (Acc@0.5) on ReferCOCO~\cite{yu2016modeling}, ReferCOCO+~\cite{yu2016modeling}, and ReferCOCOg~\cite{mao2016generation}. The results of previous best two-stage and one-stage methods are highlighted with underlines. We highlight our results in bold. The results demonstrate that our method outperforms all state-of-the-art one-stage and two-stage methods.}
	
	\small
	\begin{center}
		\scalebox{0.8}[0.8]{
			\setlength
			\tabcolsep{3.8pt}
			\begin{tabular}{c | c | c c c | c c c | c c c }
				\toprule
				\multirow{2}{*}{Models} & \multirow{2}{*}{Backbone} & \multicolumn{3}{c|}{RefCOCO} & \multicolumn{3}{c|}{RefCOCO+} & \multicolumn{3}{c}{RefCOCOg} \\ 
				
				&  & val & testA & testB & val & testA & testB & val-g & val-u & test-u \\
				\hline \hline
				\textbf{\textit{Two-stage:}} &  & & & & & & & & &\\
				CMN~\cite{hu2017modeling} & VGG16 & - & 71.03 & 65.77 & - & 54.32 & 47.76 & 57.47 & - & - \\
				VC~\cite{zhang2018grounding} & VGG16 & - & 73.33 & 67.44 & - & 58.40 & 53.18 &62.30 & - & - \\
				MAttNet~\cite{yu2018mattnet} &	ResNet-101 & 76.65 & 81.14 & 69.99 & 65.33 & 71.62 & 56.02 & -  & 66.58 & 67.27 \\
				LGRANs~\cite{wang2019neighbourhood} & VGG16 & - & 76.60 & 66.40 & - & 64.00 & 53.40 & 61.78 & - & - \\
				DGA~\cite{yang2019dynamic} & VGG16 & - & 78.42 & 65.53 & - & 69.07 & 51.99 & - & - & 63.28 \\ 
				RvG-Tree~\cite{hong2019learning} & ResNet-101 & 75.06 & 78.61 & 69.85 & 63.51 & 67.45 & 56.66 & - & 66.95 & 66.51 \\
				NMTree~\cite{liu2019learning} & ResNet-101 & 76.41 & 81.21 & 70.09 & 66.46 & 72.02 & 57.52 & 64.62 & 65.87 & 66.44 \\
				\hline
				\textbf{\textit{One-stage:}} &  & & & & & & & & &\\
				SSG~\cite{chen2018real} &  DarkNet-53 & - & 76.51 & 67.50 & - & 62.14 & 49.27 & 47.47 & 58.80 & - \\  
				FAOA~\cite{yang2019fast} & DarkNet-53 & 72.54 & 74.35 & 68.50 & 56.81 & 60.23 & 49.60 & 56.12 & 61.33 & 60.36 \\
				RCCF~\cite{liao2020real} & DLA-34 & - & 81.06 & 71.85 & - & 70.35 & 56.32 & -  & - & 65.73 \\
				ReSC-Large~\cite{yang2020improving} & DarkNet-53 & 77.63 & 80.45 & 72.30 & 63.59 & 68.36 & 56.81 & 63.12 & 67.30 & 67.20 \\
				SAFF~\cite{ye2021one}& DarkNet-53& 79.26 &81.09 &76.55 &64.43 &68.46 &58.43 &- &68.94 &68.91\\
				HFRN~\cite{qiu2020language}& ResNet-101 & 79.76 &83.12 &75.51 &66.80 &72.53 &59.09 &- &69.71 &69.08\\
				LBYL-Net~\cite{huang2021look}& DarkNet-53&  79.67 &82.91 &74.15 &68.64 &73.38 &59.49 &62.70 &- &-\\
				TransVG\cite{deng2021transvg} & ResNet-101 & 81.02 &82.72 &78.35 &64.82 &70.70 &56.94 &67.02 &68.67 &67.73 \\
                    TransVG (Swin) & Swin-S & 82.33 & 84.01 & 79.83 & 64.94 & 70.19 & 56.47 & 67.81 & 69.34 & 68.99 \\
				QRNet~\cite{ye2022shifting}& Swin-S& \underline{84.01} &\underline{85.85} &\underline{82.34} &\underline{72.94} &\underline{76.17} &\underline{63.81} &\underline{71.89} &\underline{73.03} &\underline{72.52}\\
				\hline
				EpmVG (ours)& Swin-S& $\mathbf{87.04}$& $\mathbf{89.21}$& $\mathbf{82.90}$&\textbf{76.45} &\textbf{81.16} &\textbf{68.12}&\textbf{76.15 }&\textbf{79.32} &\textbf{76.61} \\
                    \bottomrule
			\end{tabular}
		} 
	\end{center}
	\label{tab7}
        \vspace{-0.5cm}
\end{table}

\begin{figure}[h]
	\centering 
	\includegraphics[width=0.8\textwidth, angle=0]{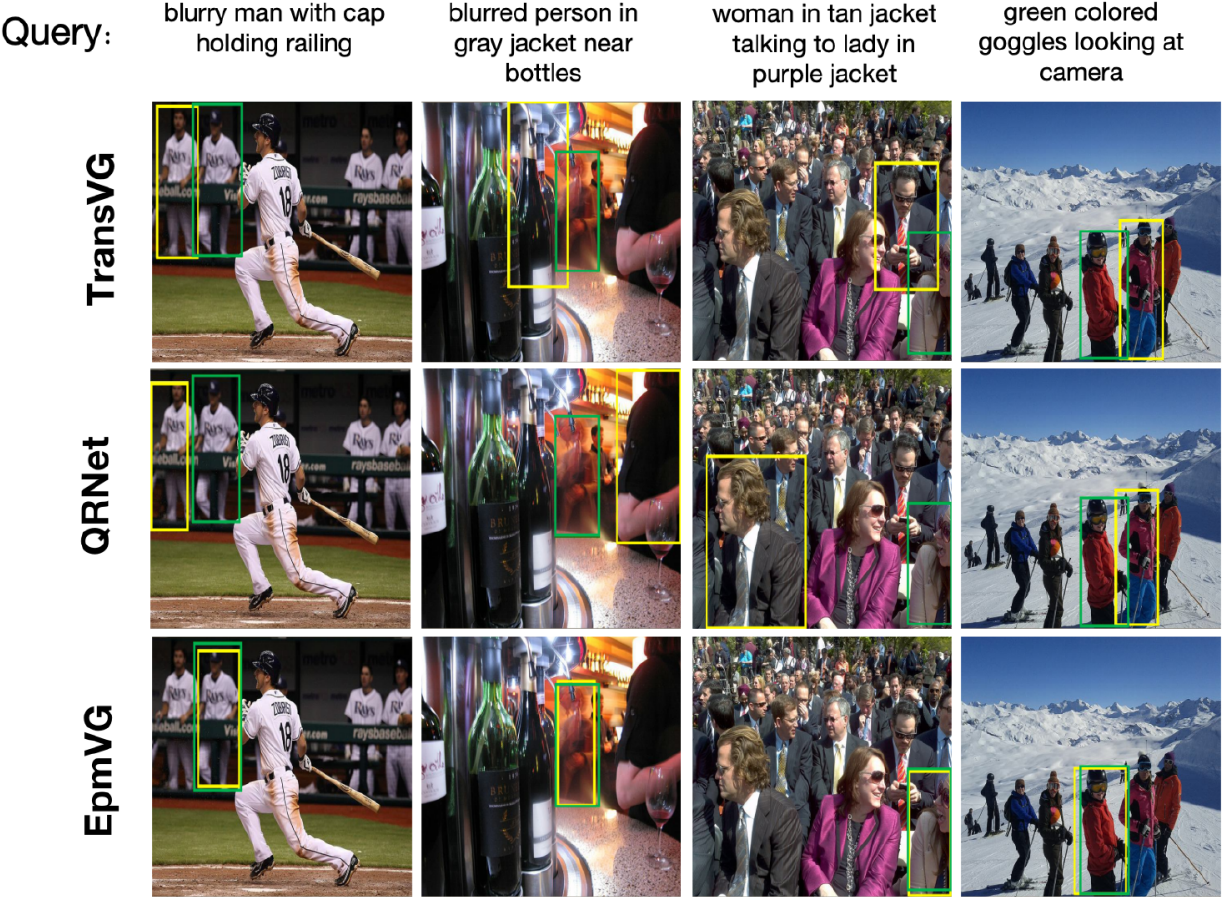}	
	\caption{Qualitative comparison with TransVG~\cite{deng2021transvg} and QRNet~\cite{ye2022shifting}. Green boxes are the ground-truth, yellow boxes are the predictions.}

	\label{fig_mom3}%
\end{figure}

We also show the accuracy comparison with state-of-the-art methods on ReferCOCO~\cite{yu2016modeling}, ReferCOCO+~\cite{yu2016modeling}, and ReferCOCOg~\cite{mao2016generation} in Table \ref{tab7}. In the ReferCOCO and ReferCOCO+ datasets, the referred objects in ``testA'' are people, and in ``testB'' they may be common objects. The expressions in ReferCOCOg are much longer than those in other datasets. According to the experimental results, our EpmVG performs excellently among all two-stage and one-stage state-of-the-art methods. On the ReferCOCO and ReferCOCO+ datasets, our method gains 3.36\%$\sim$4.99\% absolute improvement in ``testA'' and 0.56\%$\sim$4.31\% in ``testB''. In the RefCOCOg test split with longer expressions, our method is also effective, gaining 4.09\% improvement. This indicates that our proposed cross-modal distillation is very stable and effective, and also demonstrates the importance of the semantic consistency information of images and text in visual grounding tasks.

Due to the utilization of pre-trained models for distillation, their parameter size is significantly larger than that of the visual grounding model, necessitating the waiting for the inference of the pre-trained model during the training process. This results in a slower training speed after distillation compared to solely training the visual grounding model. However, during the inference process, the assistance of the pre-trained model is no longer required, therefore, our method achieves equivalent inference speed to the original visual grounding model.

\subsection{Parameter Analysis}

To investigate the influence of the hyperparameters introduced in the loss function on our method, we conducted parameter experiments on the loss weight $\lambda$ to further verify the impact of different modalities on the model performance. All parameter experiment results are obtained on RefCOCO. And the VG model was based on TransVG~\cite{deng2021transvg}, and the backbone network was ResNet-50.

\begin{table}[t]
    \caption{Effects of the loss function weights $\lambda$ on RefCOCO Dataset.}
    
    \begin{center}
	\scalebox{1.0}[1.0]{
	\setlength
	\tabcolsep{9.4pt}
           \begin{tabular}{c ccc}
           \toprule
            \multirow{2}{*}{ $\lambda$ }& \multicolumn{3}{c}{ RefCOCO } \\ 
            & val & testA & testB \\
            \hline\hline
             0& 77.81& 80.25&72.34\\ 
             0.5& 82.52& 85.16&77.82\\
 1& $\mathbf{83.39}$ & $\mathbf{86.20}$ &$\mathbf{78.98}$ \\
 1.5& 82.98& 85.56&78.03\\
             2& 82.24& 84.88&77.42\\ 
             \bottomrule
        \end{tabular}
		} 
	\end{center}
	\label{tab8}
	\vspace{-1.0cm}
\end{table}

The experimental results in Table~\ref{tab8} indicate that the GIoU loss is crucial. When $\lambda$, the model performance significantly decreases. If the weight of the GIoU loss is too small, the model primarily focuses on other loss components while neglecting precise regression of the bounding boxes. This leads to insufficient optimization of the bounding box localization, resulting in inaccurate predictions of the bounding box positions and sizes. Conversely, when $\lambda$ is too large, the model mainly focuses on bounding box regression and overlooks learning image-text consistency information from the pre-trained model. Overemphasizing the bounding box regression task may hinder the effective learning of the cross-modal distillation task, thereby decreasing overall performance.


\subsection{Qualitative Results}

We present a qualitative comparison between our EpmVG and three popular methods in Figure \ref{fig_mom3}. We observe that our method can successfully model more complex queries, such as ``woman in tan jacket talking to lady in purple jacket'' in Figure \ref{fig_mom3}. We can see that previous methods are not able to understand the correspondence between complex queries and objects in the image well, leading to incorrect predictions, such as the results of TransVG~\cite{deng2021transvg} and QRNet~\cite{ye2022shifting} for the queries of ``blurry man with cap holding railing'' and ``green colored goggles looking at camera''. In contrast, our EpmVG generates query-consistent features by our designed cross-modal distillation and makes more accurate predictions. This cross-modal distillation can introduce the image-text consistency information contained in the CLIP~\cite{radford2021learning} model, helping the model to better understand the relationship between the query and the image, thereby generating more accurate features and reducing the sensitivity to query-unrelated areas.

\section{Conclusion}
In this paper, we argue that the visual and linguistic backbone used in current visual grounding methods has a modality  domain gap since they are separately pre-trained with one single modality. To overcome this weakness, we propose EpmVG, which aims to empower the pre-trained model for visual grounding. Specifically, we propose a novel cross-modal distillation mechanism, which effectively distills the cross-modal information within the CLIP model, thus reducing the existing domain gap and improving the grounding performance. Extensive experiments indicate that the proposed framework significantly outperforms the state-of-the-art methods.

\bibliographystyle{splncs04}
\bibliography{EpmVG}

\end{document}